%% file: main.tex
\documentclass[10pt,twocolumn,letterpaper]{article}

\usepackage{wacv}

\input{preamble}

\def\wacvPaperID{2032}
\def\confName{WACV}
\def\confYear{2027}

\title{REDiff-AD: No Training and No Target Masks for Industrial Anomaly Image Generation via Diversity-Driven In-Context MMDiT}

\input{config/authors}

\newif\ifincludelatersections
\includelatersectionstrue

\begin{document}
\maketitle

\input{sections/abstract}
\input{sections/introduction}
\input{sections/related_work}
\input{sections/preliminaries}
\input{sections/method}
\ifincludelatersections
  \input{sections/experiments}

  \input{sections/conclusion}
\fi

\startreferenceslayout
\input{sections/acknowledgments}
{\small
\setlength{\bibsep}{1pt}
\bibliographystyle{ieeenat_fullname}
\bibliography{references}
}

\end{document}

%% file: preamble.tex
\input{config/packages}
\input{config/macros}

\input{config/layout}

%% file: config/packages.tex
\usepackage{amsmath,amssymb,bm}
\usepackage[accsupp]{axessibility}
\usepackage{booktabs}
\usepackage{graphicx}
\usepackage{multirow}
\usepackage{makecell}
\usepackage{xcolor}
\usepackage{placeins}
\usepackage{tikz}
\usetikzlibrary{arrows.meta,calc,fit,positioning}

\definecolor{panelpeach}{RGB}{252,238,228}
\definecolor{wacvblue}{rgb}{0.21,0.49,0.74}
\usepackage[pagebackref,breaklinks,colorlinks,allcolors=wacvblue]{hyperref}

%% file: config/macros.tex
\newcommand{\method}{REDiff-AD}

%% file: config/layout.tex
\makeatletter
\newcommand{\startreferenceslayout}{%
  \clearpage
  \global\@colht\textheight
  \global\@colroom\textheight
  \global\vsize\textheight
  \raggedbottom
}
\newcommand{\compactsubsection}[1]{%
  \@startsection{subsection}{2}{\z@}{-5pt}{1pt}{\elvbf}%
    {\texorpdfstring{\hskip -1em.~}{}#1}%
}
\makeatother

\newcommand{\compactdisplayspacing}{%
  \setlength{\abovedisplayskip}{3pt}%
  \setlength{\belowdisplayskip}{3pt}%
  \setlength{\abovedisplayshortskip}{3pt}%
  \setlength{\belowdisplayshortskip}{3pt}%
}
\AtBeginDocument{%
  \apptocmd{\normalsize}{\compactdisplayspacing}{}{}%
  \compactdisplayspacing
  \ifincludelatersections\else\raggedbottom\fi
}

\newcommand{\papergraphic}[2][]{%
  \scalebox{1}[0.85]{\includegraphics[#1]{#2}}%
}

\makeatletter
\apptocmd{\@maketitle}{%
  \vspace{-3.0em}%
  {\centering
    \scalebox{1}[0.98]{\includegraphics[width=\textwidth]{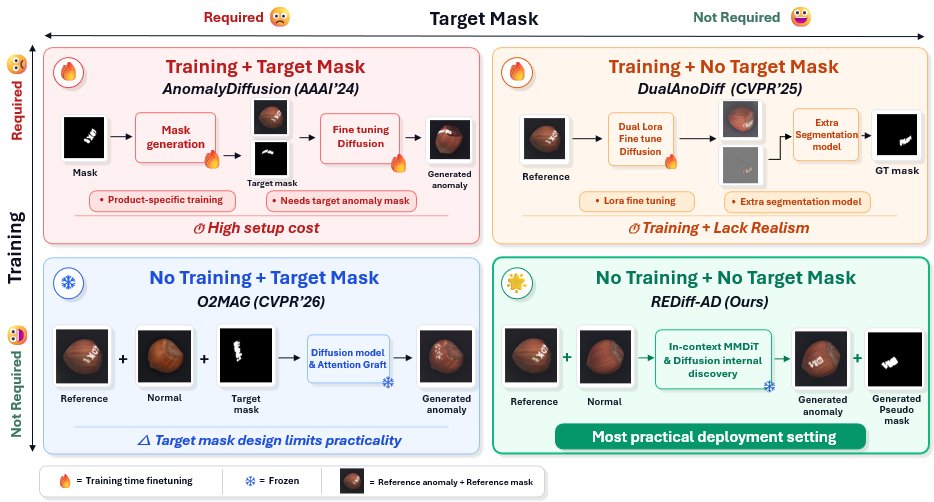}}\par
    \captionsetup{type=figure,skip=2pt}
    \captionof{figure}{Comparison of representative industrial anomaly generation approaches. REDiff-AD uses frozen in-context MMDiT and diffusion internal discovery to generate realistic anomalies using pseudo-masks but without training or target masks.}
    \label{fig:teaser}
  \par}
  \vspace{0.35em}%
}{}{}
\makeatother

%% file: config/authors.tex
\author{%
Chih-Yu Tsai\thanks{Equal contribution.}\quad
Chih-Heng Chang\footnotemark[1]\quad
Jian-Jiun Ding\\
National Taiwan University, Taipei, Taiwan\\
{\small\texttt{\{r13942126,r14942077,jjding\}@ntu.edu.tw}}%
}

\makeatletter
\hypersetup{%
  pdftitle={\@title},
  pdfauthor={Chih-Yu Tsai; Chih-Heng Chang; Jian-Jiun Ding},
  pdflang={en-US}%
}
\makeatother

%% file: sections/abstract.tex
\begin{abstract}

\vspace{-3pt}

Industrial anomaly detection is challenged by the scarcity and diversity of real defects, leaving insufficient anomalous data to effectively train downstream models. Existing anomaly generation methods often require product-specific fine-tuning, target mask annotation, or costly procedures to achieve realistic and diverse defects. We propose \method, a framework that requires neither model training nor target masks and introduces Rectified Flow MMDiT into industrial anomaly generation for efficient sampling with fewer generation steps. Given a normal product image and a single masked reference anomaly, \method{} employs in-context learning to transfer the defect appearance from the reference into the normal image. Internal attention signals generated during this process are further reused to localize the synthesized anomaly, while edge and color cues refine the localization into pixel-level pseudo-masks. Two lightweight modules further improve generation quality: adaptive reference conditioning enhances defect fidelity, while diversity guidance promotes varied defect patterns without additional training. Experiments demonstrate that \method{} can well address the tradeoff between realism and diversity for anomaly pattern generation and achieve competitive downstream anomaly classification, detection, and localization performance. These results show that \method{} is a simple, efficient, and effective solution for large-scale anomaly data augmentation and pixel-level data construction. Github code:
\url{https://github.com/efdefe4-droid/REDiff-AD}.

\end{abstract}

\vspace{-9pt}

%% file: sections/introduction.tex
\section{Introduction}
\label{sec:introduction}

Industrial product inspection relies heavily on sufficient and diverse training data for anomaly classification, detection, and localization. In real manufacturing environments, however, normal products are abundant, whereas real defects are rare, resulting in severely imbalanced training sets and insufficient coverage of different defect patterns~\cite{bergmann2019mvtec}. Such data imbalance can substantially degrade the learning of rare categories, as widely observed in anomaly classification and detection~\cite{cao2019learning,tan2020equalization}. The problem becomes even more challenging for pixel-level tasks, where accurately annotating irregular defect boundaries requires substantial manual effort. Therefore, generating realistic and diverse anomaly images together with reliable pixel-level masks provides a practical way to enrich scarce anomalous samples, balance the training data, and provide sufficient supervision for downstream industrial inspection models.

Early industrial anomaly generation methods were mainly based on generative adversarial networks (GANs)~\cite{goodfellow2014generative}, which typically require learning the target industrial distribution from task-specific data~\cite{niu2020defect,zhang2021defectgan,duan2023dfmgan}. However, adversarial training is difficult to stabilize when defect samples are limited, and learning complex defect appearance from scarce anomalies can lead to distorted or insufficiently diverse generations while requiring substantial training effort. Pretrained diffusion models offer a stronger alternative by providing rich visual priors learned from large-scale image collections~\cite{rombach2022high}. However, generic diffusion priors do not directly capture the fine-grained appearance of specific industrial products and defects, creating a domain gap between generic synthesis and real production data. Consequently, recent methods adapt pretrained diffusion models to industrial domains~\cite{hu2024anomalydiffusion,jin2025dualanodiff,dai2025seas,song2025defectfill}, often building on personalization techniques such as Textual Inversion, Custom Diffusion, and LoRA~\cite{gal2023textualinversion,kumari2023customdiffusion,hu2022lora}. Although these approaches improve the fidelity of defect detection, each new product or defect typically requires additional optimization, computation, and model storage, and the problem becomes more severe in the one-shot setting where a single anomaly provides insufficient supervision for reliable adaptation.

To remove this repeated training cost, recent approaches ~\cite{xu2025tfidg,rao2026o2mag} exploit pretrained image-editing priors, such as AnyDoor and MasaCtrl~\cite{chen2024anydoor,cao2023masactrl}, for one-shot anomaly transfer. These methods avoid product-specific model training, but their reference transfer still relies on a predefined target-side mask to determine where and how the defect should be synthesized. Precise target-side masks are difficult to prepare for thin and irregular anomalies. These masks require additional annotation and training costs. An alternative direction removes the real anomaly reference entirely and generates defects from textual semantics, as explored by AnomalyAny and AnoStyler~\cite{sun2025anomalyany,so2026anostyler}. However, without observing a real defect, text-driven generation cannot directly preserve its fine-grained color, texture, and local structure, making high-fidelity industrial anomaly synthesis more challenging. A practical industrial anomaly generator should have small costs for data preparation and computation while producing realistic and diverse anomalies without repeated model training or manually specified target-side masks.

\begin{figure}[!t]
  \centering
  \papergraphic[width=0.84\linewidth]{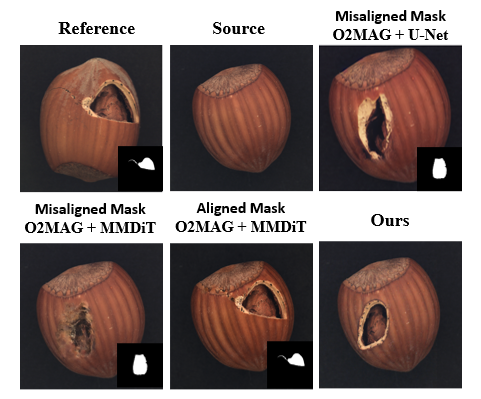}
  \caption{O2MAG handles mask misalignment on U-Net (top right), but its RoPE-based MMDiT graft fails when misaligned (bottom left) and works well only when aligned (bottom middle)~\cite{rao2026o2mag,wei2025freeflux}. Our REDiff-AD approach is the first target-mask-free industrial anomaly editor on MMDiT and it works well even when there is no aligned mask (bottom right).}
  \label{fig:o2mag-mmdit-failure}
\end{figure}

Building on in-context learning from NLP~\cite{brown2020language}, we propose \method{}, which transfers the in-context image-editing capability of Insert Anything~\cite{song2026insert} to industrial anomaly generation. We formulate one-shot anomaly synthesis as visual in-context editing: given a single masked reference anomaly and an unlabeled normal image, a frozen FLUX.1 Fill MMDiT~\cite{bfl2024fill,esser2024scaling} transfers defect appearance without product/defect-specific optimization or manually specified target masks. Its Flow Matching backbone~\cite{lipman2023flow,liu2023flow} enables efficient sampling with fewer steps. Motivated by the spatial information carried by MMDiT attention~\cite{kim2025seg4diff}, \method{} further reuses internal target-to-reference interactions for anomaly localization and generation feedback. This enables aligned pixel-level pseudo-mask construction, robust small-defect formation, and diverse anomaly generation without using an external mask generator.

The main contributions of this work are:
\begin{itemize}
  \item We introduce an industrial anomaly generation approach that repurposes in-context image editing instead of adapting diffusion models to individual products or defects. Given an anomaly reference, a frozen Flow Matching MMDiT generates reference-guided defects with fewer sampling steps than state-of-the-art diffusion-based anomaly generators, requiring neither product- or defect-specific training nor manually specified target masks.

  \item We develop a target-mask-free paired-annotation pipeline through a systematic analysis of FLUX.1 Fill. Target-to-reference attention from selected expert single-stream blocks provides accurate coarse anomaly localization. Defect-aware edges and color cues are applied to refine pixel-level pseudo-masks for downstream supervision without an external mask generator.

  \item We identify and address two failure modes of in-context anomaly generation. AAM uses stepwise attention feedback to recover weak or small defects, while DOK reduces reference-shape copying in the key space while preserving the appearance carried by Value features, improving generation robustness and diversity through lightweight and training-free operations.
\end{itemize}

%% file: sections/related_work.tex
\section{Related Work}
\label{sec:related-work}

\subsubsection*{Industrial Anomaly Generation}
Industrial anomaly synthesis can be performed by task-specific GANs to pretrained diffusion priors\cite{niu2020defect,zhang2021defectgan,duan2023dfmgan}. Training-based methods, including AnomalyDiffusion, DualAnoDiff, SeaS, and DefectFill~\cite{hu2024anomalydiffusion,jin2025dualanodiff,dai2025seas,song2025defectfill}, require product- or defect-specific optimization and model maintenance. Adaptation-free methods such as TF-IDG and O2MAG~\cite{xu2025tfidg,rao2026o2mag} avoid diffusion-weight updates but still require predefined target-side masks. It also optimizes a reference-specific text embedding process. Because it makes attention position-dependent~\cite{wei2025freeflux}, O2MAG's attention grafting cannot be directly applied to MMDiT. As in Fig.~\ref{fig:o2mag-mmdit-failure}, a misaligned graft fails, whereas editing succeeds when the target and reference masks are geometrically aligned. Instead, the proposed REDiff-AD model works directly with MMDiT, producing more accurate pixel-level masks with fewer sampling steps. Zero-shot methods such as AnomalyAny and AnoStyler~\cite{sun2025anomalyany,so2026anostyler} eliminate a real anomaly reference but forgo direct access to its fine-grained appearance. REDiff-AD retains one masked real anomaly while requiring neither product-/defect-specific adaptation nor a target-side anomaly contour.

\subsubsection*{Diffusion Internal Representation Discovery}
Pretrained diffusion models provide spatial representations of patterns and can be repurposed without training a separate segmentor. DiffSeg~\cite{tian2024diffseg} aggregates multi-resolution self-attention from U-Net-based Stable Diffusion and merges similar maps into semantic masks. Seg4Diff~\cite{kim2025seg4diff} instead identifies an expert image-to-text layer in SD3's MMDiT. In ~\cite{kim2025seg4diff}, they showed that  MMDiT-based Seg4Diff achieves higher mIoU than U-Net-based DiffSeg, while its shared token grid avoids cross-resolution resizing before aggregation. Both methods localize semantic objects within a single image. Since the target-mask ground truth is unavailable in our setting, the proposed REDiff-AD model extracts internal target-to-reference attention as a coarse anomaly mask. Compared with U-Net attention, MMDiT provides more accurate localization and simplifies mask construction through its shared token grid, but requires prior model analysis to identify the relevant expert layers.

\subsubsection*{Diversity and Representation Steering}
Strong reference conditioning in in-context generation often causes repeated copying and limits output diversity. Existing methods increase variation by relaxing such dependence or directly steering generative representations. Particle Guidance~\cite{corso2024particle} couples multiple trajectories through repulsive potentials, but its joint sampling becomes costly for large-scale dataset generation. CADS~\cite{sadat2024cads} gradually relaxes the conditioning, but weakening the condition can also suppress reference-specific defect characteristics. BoundaryDiffusion~\cite{zhu2023boundary} and AnchorSteer~\cite{chang2026anchorsteer} manipulate intermediate representations for semantic editing, but do not explicitly preserve personalized appearance during diversification. The proposed REDiff-AD model instead separates these roles within attention: we relax reference dependence through the key features that govern retrieval geometry, while preserving the features about defect appearance. This enables diverse defect shapes without discarding the reference-specific anomaly characteristics.

%% file: sections/preliminaries.tex
\section{Preliminaries}
\label{sec:preliminaries}

\subsection{Flow Matching and Rectified Flow}
\label{sec:flow-matching}

Score-based diffusion repeatedly evaluates a score network while solving a reverse-time SDE or a probability-flow ODE~\cite{song2021scorebased}, whereas Flow Matching learns an ODE velocity field~\cite{lipman2023flow}. Given noise $\epsilon\sim\mathcal N(0,I)$ and data $x_0$, Rectified Flow follows the straight path~\cite{liu2023flow}:
\begin{equation}
  x_t=(1-t)\epsilon+t x_0, \qquad t\in[0,1].
  \label{eq:rectified-flow}
\end{equation}
The resulting near-straight trajectory reduces the discretization error and often requires fewer solving steps than iterative score-based sampling~\cite{liu2023flow}.

\subsection{Attention in MMDiT}
\label{sec:mmdit-attention}

Given tokens $X\in\mathbb R^{N\times d_m}$, attention projects $Q=XW_Q$, $K=XW_K$, and $V_{\mathrm{val}}=XW_V$, the attention is computed from~\cite{vaswani2017attention}
\begin{equation}
  A=\operatorname{softmax}\!\left(\frac{QK^\top}{\sqrt d}\right),
  \qquad O=AV_{\mathrm{val}}.
  \label{eq:attention-basics}
\end{equation}
Each entry $A_{ij}$ weights retrieval from key $j$ by query $i$ with head dimension $d$. MMDiT jointly attends over modality-specific image and conditioning projections~\cite{esser2024scaling}. For target and reference indices $T$ and $R$, $A[T,R]$ is our localization readout. Keys determine the affinities, whereas Values carry the content. Thus, our Key-only operation changes retrieval geometry without directly modifying defect appearance.

%% file: sections/method.tex
\begin{figure*}[!t]
  \centering
  \papergraphic[width=\textwidth]{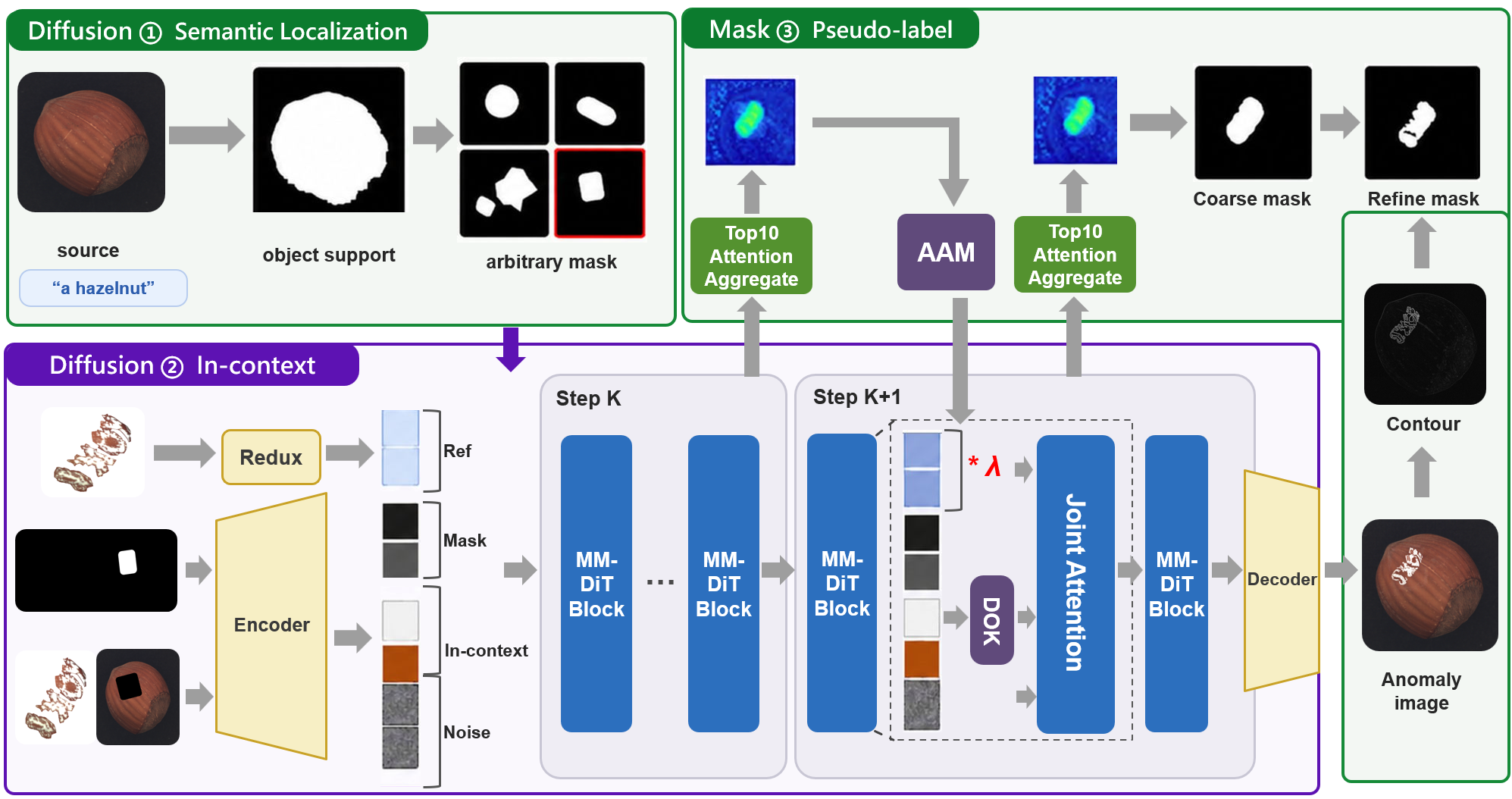}
  \caption{Overview of the proposed \method{} model. Semantic localization first determines a loose editable region on the target. A frozen in-context MMDiT transfers the masked reference, with AAM strengthening weak anomalies and DOK reducing reference-shape copying. Aggregated target-to-reference attention is then refined into the output pseudo-mask.}
  \label{fig:pipeline}
\end{figure*}

\section{Methodology}
\label{sec:method}

We propose \method{}, a one-shot industrial anomaly generation framework that requires neither model training nor manually specified target masks. The name \method{} reflects the repeated use of diffusion throughout the framework, while ``RED'' evokes the red color commonly used to highlight anomalous regions. The overall framework is illustrated in Fig.~\ref{fig:pipeline}. The remainder of this section illustrates the framework in four parts. Sec.~\ref{sec:in-context-generation} describes how in-context generation enables one-shot defect transfer using neither product-specific training nor a manually provided target mask. Sec.~\ref{sec:dok} introduces the lightweight diversity orthogonal K (DOK) module to address the diversity limitation of in-context anomaly generation. Sec.~\ref{sec:attention-pseudo-mask} analyzes the internal attention of FLUX.1 Fill and refines the resulting localization into pixel-level pseudo-masks. Finally, Sec.~\ref{sec:aam} presents the attention adaptive module (AAM), which reuses the attention signals from Sec.~\ref{sec:attention-pseudo-mask} as feedback to improve the robustness of small-defect generation.

\subsection{In-Context One-Shot Generation}
\label{sec:in-context-generation}

FLUX.1 Fill is a Rectified Flow MMDiT model for prompt driven image inpainting~\cite{bfl2024fill}. However, text prompts alone cannot reliably describe the fine grained appearance of real industrial defects, often leading to a gap from the target product domain. We therefore adopt Insert Anything~\cite{song2026insert} to introduce visual in-context conditioning, allowing a single real reference anomaly to directly guide defect generation.

Given a reference anomaly mask pair $(x^r,m^r)$ and a normal target image $x^n$, the reference defect is placed on the left and the masked target image on the right:
\begin{equation}
  \begin{aligned}
    x^{\mathrm{ctx}}
    &= \left[ R(x^r,m^r) \mid x^n \odot (1-M_{\mathrm{plan}}) \right], \\
    M^{\mathrm{ctx}}
    &= \left[ 0 \mid M_{\mathrm{plan}} \right],
  \end{aligned}
\end{equation}
where $R(\cdot)$ extracts the anomaly and $M_{\mathrm{plan}}$ specifies the edited region. Insert Anything encodes the reference image to obtain visual semantic tokens $X_C$ and jointly processes it with the reference and target image tokens:
\begin{equation}
  X_{\mathrm{joint}}=[X_C;X_R;X_T],
\end{equation}
where $X_C$ provides the semantic guidance of the reference image, while joint attention allows target tokens $X_T$ to retrieve the defect information from the reference tokens $X_R$. This enables the real reference defect to guide generation through in-context conditioning rather than relying only on text.

Unlike existing anomaly transfer methods that require a precise target mask to prescribe the final defect shape, REDiff-AD only requires a simple planning mask indicating \emph{where} editing is allowed, which is automatically generated within the product region as described in Sec.~\ref{sec:attention-pseudo-mask}. Combining the inpainting capability of FLUX.1 Fill with the visual conditioning of Insert Anything, the reference defect can be transferred to this region without product- or defect-specific fine tuning. However, anomaly generation requires diverse samples rather than exact reproduction of the reference defect. We therefore introduce DOK in Sec.~\ref{sec:dok} to improve the diversity of in-context anomaly generation.

\begin{figure}[!t]
  \centering
  \papergraphic[width=\columnwidth]{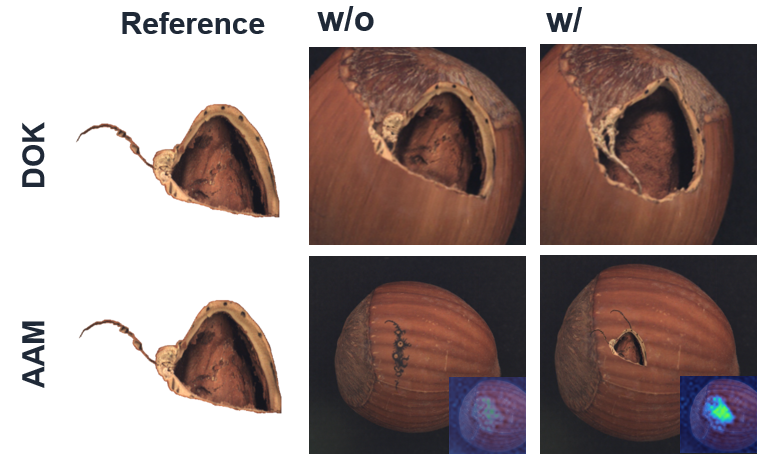}
  \caption{Qualitative illustration of the effects of DOK and AAM. DOK diversifies the geometry of the defect, while AAM improves the robustness of in-context defect generation.}
  \label{fig:dok-aam}
\end{figure}

\subsection{Diversity Orthogonal K (DOK)}
\label{sec:dok}

In-context generation strongly preserves the visual reference, which is beneficial for personalized editing but undesirable for anomaly augmentation, where diverse defect shapes are required. As shown in Fig.~\ref{fig:dok-aam}, direct in-context generation tends to reproduce the geometry of the reference defect, which limits the diversity. Existing diversity-oriented methods either incur substantial computational overhead or over-edit the input, failing to preserve product-specific defect characteristics. We therefore introduce Diversity Orthogonal K (DOK), a lightweight operation that relaxes reference shape dependence while preserving defect appearance.

Given the reference mask $m^r$, we resize it to the reference token grid and flatten it to $s\in\mathbb{R}^{N}$. Its spatial shape direction is represented as
\begin{equation}
  e=
  \frac{s-\bar{s}\mathbf{1}}
  {\lVert s-\bar{s}\mathbf{1}\rVert_2},
\end{equation}
where $\bar{s}$ denotes the mean of $s$. For each reference Key feature $K_R$, we suppress the component aligned with the same direction:
\begin{equation}
  \widetilde{K}_R=K_R-\eta e e^{\top}K_R,
\end{equation}
where $\eta$ controls the relaxation strength. In contrast, the reference Value features remain unchanged,
\begin{equation}
  \widetilde{V}_R=V_R.
\end{equation}

Since Keys determine how target tokens retrieve the reference information, reducing the reference-shape component relaxes the geometric constraint imposed by in-context conditioning. Meanwhile, preserving $V_R$ retains the defect appearance. As illustrated in Fig.~\ref{fig:dok-aam}, DOK therefore allows the generated anomaly to deviate from the reference geometry while maintaining its characteristic visual features. The operation is applied only during intermediate flow steps and requires no additional training or auxiliary model.

\begin{figure}[!t]
  \centering
  \papergraphic[width=0.95\columnwidth]{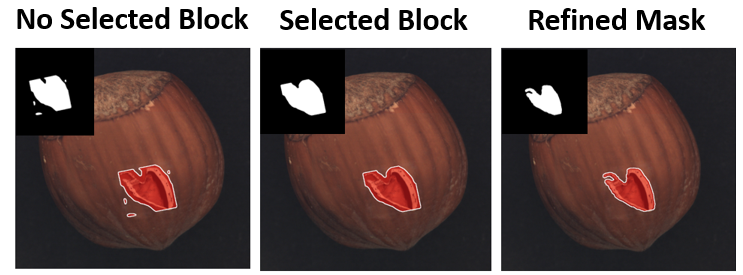}
  \caption{Block selection and pixel-level refinement. Red overlays show localized anomalies. Selected blocks produce a cleaner coarse mask, while refinement improves pixel-level accuracy.}
  \label{fig:mask-comparison}
\end{figure}

\subsection{Target-Mask-Free Pseudo-Mask Generation}
\label{sec:attention-pseudo-mask}

\noindent\textbf{Planning region.}\ 
For the methods that require target masks, it is necessary to provide precise defect annotations and carefully paired source--target images together with the defect located within the region of the source object. To avoid this requirement, REDiff-AD uses diffusion-based localization to identify the target object and constrains defect generation to its interior. Specifically, we first invert the target image using the Rectified Flow solver (RF-Solver)~\cite{wang2025rfsolver}. We then aggregate image-to-text attention during inversion to obtain the object support $\Omega$, within which the defect region is generated. One or two simple geometric primitives $P_j$ are randomly sampled inside the product support:
\begin{equation}
  M_{\mathrm{plan}}
  =
  \bigcup_{j=1}^{N}
  (P_j\cap\Omega),
  \qquad
  N\in\{1,2\}.
\end{equation}
Thus, $M_{\mathrm{plan}}$ provides an approximate defect location by randomly placing simple shapes within the product support, rather than specifying the final defect contour.

\noindent\textbf{Attention-based coarse localization.}\ 
Previous studies show that internal diffusion attention contains useful spatial semantics~\cite{tian2024diffseg,kim2025seg4diff}. However, FLUX.1 Fill additionally contains reference and target image tokens, making it unclear which attention direction and blocks best describe the transferred anomaly. We therefore compare target-to-target (T2T) and target-to-reference (T2R) responses over all 57 FLUX.1 Fill blocks. For the block $b$ and the flow step $t$, the T2R response is
\begin{equation}
  A_{\mathrm{T2R}}^{b,t}
  =
  \mathrm{Softmax}
  \left(
    \frac{Q_TK_R^{\top}}{\sqrt{d}}
  \right).
\end{equation}

\begin{figure}[!t]
  \centering
  \papergraphic[width=0.82\columnwidth]{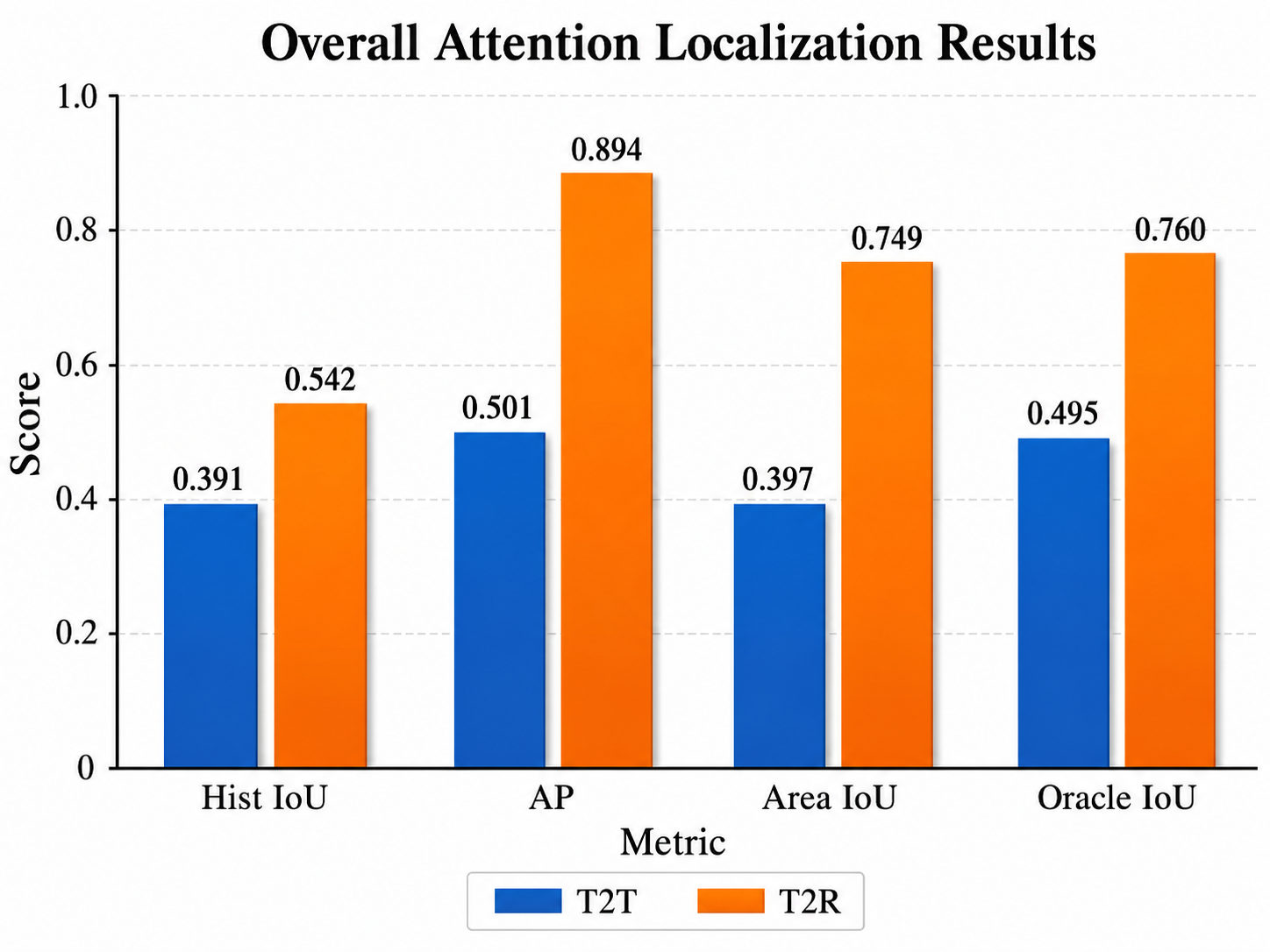}
  \caption{T2T and T2R attention localization. T2R achieves higher scores across all metrics.}
  \label{fig:t2t-t2r}
\end{figure}

We then perform block analysis using 30 anomaly samples and 41 sampled flow steps. Each block is aggregated over steps, thresholded, and ranked by localization IoU. The most frequently selected blocks form a fixed top-10 expert set used for subsequent generation. We find that T2R provides clearer defect--background separation than T2T and nine of the ten selected blocks belong to the single-stream stage. As shown in Fig.~\ref{fig:t2t-t2r}, T2R achieves higher localization scores than T2T across all evaluated metrics. Their aggregated response produces the coarse anomaly mask $M_c$. As shown in Fig.~\ref{fig:mask-comparison}, expert-block selection suppresses scattered responses and produces a cleaner coarse mask, which is important because a reliable coarse region substantially reduces irrelevant responses during subsequent pixel-level refinement.

\noindent\textbf{Pixel-level refinement.}\ 
Industrial defects are often small, making token-level attention insufficient for recovering precise boundaries~\cite{chen2026bocchirealisticchallengingbenchmark}. Nevertheless, anomalous regions commonly introduce local high-frequency structures and appearance changes that differ from the surrounding normal surface. We therefore use $M_c$ to define the refinement region and recover the pixel-level information directly from the generated image.
We first combine Scharr gradients and Haar high-frequency responses to capture defect boundaries and fine structures:
\begin{equation}
  \begin{aligned}
    E
    &= \mathrm{Norm}
    \left(
      0.55|\nabla_{\mathrm{Scharr}}g|
      +
      0.45|\mathrm{HHF}(g)|
    \right), \\
    Q
    &= \left[
      E>P_{80}(E\mid M_c)
    \right],
  \end{aligned}
\end{equation}
where $g$ denotes the grayscale generated image and $Q$ represents high-confidence structural responses. Since some defects have weak edges but exhibit different colors or appearances from the normal product surface, we further compare foreground and surrounding background statistics in the CIELAB space:
\begin{equation}
  M_{\mathrm{Lab}}(p)
  =
  \mathbf{1}
  \left[
    d_f(p)<(1+\delta)d_b(p)
  \right],
\end{equation}
where $d_f(p)$ and $d_b(p)$ denote the appearance distances of the pixel $p$ to the estimated defect and normal regions, respectively. The structural response $Q$ and the appearance response $M_{\mathrm{Lab}}$ are combined within $M_c$ and followed by simple morphological refinement to obtain the final pixel-level pseudo mask. As illustrated in Fig.~\ref{fig:mask-comparison}, the clean attention-based coarse mask provides a reliable starting region, while image-level cues recover the finer defect boundary required for downstream supervision.

\subsection{Attention Adaptive Module (AAM)}
\label{sec:aam}
In-context anomaly transfer may fail for small or weak defects, as in Fig.~\ref{fig:dok-aam}. Simply increasing the reference condition is not sufficient: Weak conditioning may fail to generate the defect, while overly strong conditioning can distort the image~\cite{lee2026rl,tsai2025adaptivewavelet}. Existing adaptive methods such as AnomalyDiffusion~\cite{hu2024anomalydiffusion} require decoding intermediate results to estimate the defect, introducing additional computation. Instead, REDiff-AD directly reuses the T2R attention from Sec.~\ref{sec:attention-pseudo-mask} as a monitoring signal and introduces a lightweight attention adaptive module (AAM) without additional forward passes or decoding.

Let $S_t$ denote the aggregated attention map at the flow step $t$. We measure the anomaly response inside and outside the planning region and define the formation score as
\begin{equation}
  s_t=
  \left(
    \mu_{\mathrm{in}},
    \mu_{\mathrm{in}}/\mu_{\mathrm{out}},
    \mu_{\mathrm{in}}-\mu_{\mathrm{out}}
  \right),
\end{equation}
where $\mu_{\mathrm{in}}$ and $\mu_{\mathrm{out}}$ are the mean attention responses inside and outside the region, respectively. A low $s_t$ increases $\lambda_t$ to strengthen conditioning, while a high $s_t$ reduces $\lambda_t$ to avoid image distortion.

The AAM adaptively scales the Redux condition tokens:
\begin{equation}
  \widetilde{X}_C^{t}=\lambda_t X_C.
\end{equation}
We enhance the Redux condition rather than the in-context reference tokens because Redux provides the semantic guidance of \emph{what} should be generated. By contrast, directly strengthening reference tokens may further reinforce its geometry and reduce the diversity. By reusing attention already computed during generation, the AAM improves the success and robustness of small-defect synthesis with negligible additional cost.

%% file: sections/experiments.tex
\section{Experiments}
\label{sec:experiments}

\input{sections/experiment_tables}

\subsection{Experimental Settings}
\label{sec:experimental-settings}
\noindent\textbf{Dataset and protocol.}\ 
We evaluate REDiff-AD on all 15 categories of MVTec-AD~\cite{bergmann2019mvtec}, covering both industrial objects and texture. For each defect type, the first one third of the real anomaly images is used as the reference or training pool, while the remaining two thirds are reserved for evaluation. REDiff-AD uses a single anomaly image--mask pair as reference and generates $250$ samples per defect type, matching the normal-data scale for balanced downstream training. All images are generated at $512\times512$ resolution using $30$ Flow Matching steps. For fair comparison, all methods use the same data split, synthetic sample budget, and downstream evaluation protocol.

\noindent\textbf{Generation quality and diversity}. The kernel inception distance (KID)~\cite{binkowski2018mmd} is applied to evaluate the generated-to-real distributional distance (lower is better) and Intra-Cluster LPIPS (IC-LPIPS)~\cite{zhang2018lpips,ojha2021fewshot} is adopted to evaluate the within-defect perceptual diversity (higher is better).

\noindent\textbf{Anomaly classification}. It evaluates whether synthetic defects provide useful discriminative features for downstream learning. We trained an ImageNet-pretrained ResNet-34~\cite{he2016resnet,deng2009imagenet} with the generated anomaly data and reported classification accuracy on held-out real anomalies.

\noindent\textbf{Anomaly detection and localization}. They evaluate the quality of the generated image--mask pairs. A discriminative U-Net~\cite{ronneberger2015unet} is trained with synthetic supervision and evaluated on real data using the AUROC~\cite{fawcett2006introduction}, Average Precision (AP)~\cite{davis2006relationship}, and the F1 score. We report both image-level metrics AP-I for anomaly detection and pixel-level metrics (AUROC-P, AP-P, F1-P) for anomaly localization.

\subsection{Comparison in Anomaly Generation}
\label{sec:comparison-anomaly-generation}
\placegenerationtables
\noindent\textbf{Baselines.}\ 
We compare REDiff-AD with representative anomaly generation methods. AnomalyDiffusion~\cite{hu2024anomalydiffusion} and DualAnoDiff~\cite{jin2025dualanodiff} are representative anomaly generation methods that adopt training-based approaches, while TF-IDG~\cite{xu2025tfidg} and O2MAG~\cite{rao2026o2mag} proposed in 2025 and 2026, respectively, are training-free reference-based methods. For both methods, we use masks generated by the fine-tuned AnomalyDiffusion model as target masks. We further evaluate the zero-shot capability of REDiff-AD against the AnoStyler~\cite{so2026anostyler} model proposed in 2026. 

\noindent\textbf{Generation quality and diversity.}\ 
Table~\ref{tab:generation-quality-diversity} quantitatively compares generation realism and diversity. REDiff-AD achieves the lowest average KID of 31.09 while maintaining competitive IC-LPIPS, demonstrating a strong realism--diversity balance. Although DualAnoDiff obtains the highest diversity, its much larger KID indicates that this variation often comes at the cost of realism. This is also reflected in Fig.~\ref{fig:qualitative-comparison}: Training-based methods can introduce noticeable appearance distortion, particularly on texture categories, while target-mask-dependent training-free methods may suffer from spatial mismatch and place defects outside the product region, as observed on the metal nut. In contrast, REDiff-AD generates diverse defects while preserving the target product appearance.

\noindent\textbf{Classification.}\ 
Table~\ref{tab:classification} shows that REDiff-AD generates diverse anomalies that can be used as effective training data for downstream classification. Although its average accuracy is slightly below O2MAG, REDiff-AD achieves the best performance on several categories. More importantly, these results are obtained without manually specified target masks, making REDiff-AD more suitable for practical industrial applications.

\noindent\textbf{Detection and localization.} \ 
Table~\ref{tab:detection} shows that REDiff-AD achieves the best average performance among training-free methods for downstream detection and localization and obtains the highest AP-I overall, indicating reliable formation of anomaly patterns at the image level. As illustrated in Fig.~\ref{fig:qualitative}, target-mask-dependent transfer can fail for small defect regions, such as O2MAG on wood, which further degrades pixel-level localization. Although AnomalyDiffusion performs better on some pixel-level metrics, REDiff-AD remains competitive without fine tuning or precise target masks.

\placeunettable

\compactsubsection{Zero-Shot Anomaly Generation}
\label{sec:zero-shot-anomaly-generation}
We further evaluate the zero-shot capability and convenience of REDiff-AD by replacing the real anomaly reference with an ordinary daily life image. As shown in Fig.~\ref{fig:zero-shot-qualitative}, a single visual reference is sufficient to generate diverse anomalies without additional training or target masks when using the proposed REDiff-AD model. Compared with the prompt-driven AnoStyler~\cite{so2026anostyler}, REDiff-AD achieves better generation fidelity and diversity, as in Table~\ref{tab:zero-shot-comparison}. The generated image--mask pairs also provide stronger supervision for downstream anomaly detection and pixel-level localization, achieving better AP-I, AUC-P, AP-P, and F1-P.

\compactsubsection{Ablation Study}
\label{sec:ablation-study}
\noindent\textbf{Runtime and memory cost.}\ 
Table~\ref{tab:efficiency} reports per-image time (by NVIDIA RTX~5090), the peak VRAM, and the Hazelnut KID. REDiff-AD takes 10\,s at 20 steps (KID 9.26), while 40 steps reaches 8.53. Its 9\,GB peak VRAM is lower than that of TF-IDG (22\,GB) and O2MAG (16\,GB). The results verify that REDiff-AD is able to generate large-scale anomaly pattern using consumer-grade GPUs\\[-1pt]
\textbf{Component experiments.}\ 
Across MVTec-AD, Table~\ref{tab:mask-ablation} shows that the planning ROI alone is insufficient as pixel-level ground truth. Selecting informative attention blocks improves localization over aggregating all blocks, while pixel-level refinement provides the largest gain, increasing AP-P and F1-P to 77.82 and 74.05, respectively. Table~\ref{tab:component-ablation} shows that in-context generation provides reasonable realism but remains limited in diversity and downstream utility. AAM improves generation robustness, increasing classification accuracy and AP-I, while DOK increases IC-LPIPS and makes downstream models able to be trained with diverse defect patterns. Together, they yield a strong realism--diversity balance and the best downstream performance, reaching 70.77 accuracy and 99.4 AP-I.

%% file: sections/experiment_tables.tex

\newcommand{\tableoneavgvalue}[1]{{\fontsize{6.2pt}{6.7pt}\selectfont #1}}
\newcommand{\tablefiveheader}[1]{{\fontsize{7.8pt}{8.3pt}\selectfont #1}}

\newcommand{\placegenerationtables}{%
\begin{table*}[!t]
\begin{minipage}[t]{0.48\textwidth}
  \vspace{0pt}
  \centering
  \captionof{table}{Generation quality and diversity on MVTec-AD.}
  \label{tab:generation-quality-diversity}
  \fontsize{5.6pt}{6.7pt}\selectfont
  \setlength{\tabcolsep}{0.5pt}
  \renewcommand{\arraystretch}{1.03}
  \begin{tabular*}{\columnwidth}{@{\extracolsep{\fill}}l|cc|cc|cc|cc|cc@{}}
    \toprule
    \multirow{2}{*}{Category}
      & \multicolumn{2}{c|}{AnoDiff}
      & \multicolumn{2}{c|}{DualAno}
      & \multicolumn{2}{c|}{TF-IDG}
      & \multicolumn{2}{c|}{O2MAG}
      & \multicolumn{2}{c}{Ours} \\
    \cline{2-11}
      & KID$\downarrow$ & IC-L$\uparrow$
      & KID$\downarrow$ & IC-L$\uparrow$
      & KID$\downarrow$ & IC-L$\uparrow$
      & KID$\downarrow$ & IC-L$\uparrow$
      & KID$\downarrow$ & IC-L$\uparrow$ \\
    \midrule
    bottle
      & 118.01 & \underline{0.17} & 214.32 & \textbf{0.39}
      & \textbf{33.25} & 0.16 & 42.52 & 0.16
      & \underline{38.66} & \underline{0.17} \\
    cable
      & 91.26 & 0.40 & 52.87 & \textbf{0.49}
      & 47.27 & 0.38 & \textbf{26.73} & \underline{0.41}
      & \underline{36.81} & 0.39 \\
    capsule
      & 113.01 & 0.18 & 62.33 & \textbf{0.32}
      & \textbf{11.76} & 0.16 & 36.43 & 0.18
      & \underline{14.44} & \underline{0.19} \\
    carpet
      & 106.99 & 0.21 & 49.26 & \textbf{0.31}
      & \underline{22.22} & 0.24 & 51.96 & 0.24
      & \textbf{18.62} & \underline{0.25} \\
    grid
      & 52.53 & \textbf{0.45} & 253.06 & \underline{0.43}
      & \underline{7.79} & 0.36 & \textbf{5.92} & 0.38
      & 8.16 & 0.36 \\
    hazelnut
      & 58.83 & 0.31 & 33.01 & \textbf{0.40}
      & \textbf{8.62} & 0.30 & 9.74 & 0.30
      & \underline{8.87} & \underline{0.32} \\
    leather
      & 219.71 & \textbf{0.39} & 203.17 & \underline{0.37}
      & \textbf{32.62} & 0.30 & 152.78 & 0.32
      & \underline{52.57} & 0.32 \\
    metal\_nut
      & 102.49 & 0.26 & 53.60 & \textbf{0.36}
      & 46.74 & 0.25 & \underline{42.96} & 0.27
      & \textbf{35.09} & \underline{0.28} \\
    pill
      & 66.29 & 0.23 & 118.63 & \textbf{0.45}
      & \underline{26.59} & 0.24 & 41.76 & 0.24
      & \textbf{20.84} & \underline{0.27} \\
    screw
      & 10.13 & 0.30 & 21.22 & \textbf{0.39}
      & 1.59 & 0.30 & \textbf{1.26} & 0.32
      & \underline{1.34} & \underline{0.33} \\
    tile
      & 182.51 & \textbf{0.49} & \underline{115.74} & 0.43
      & 118.43 & 0.45 & 124.84 & 0.46
      & \textbf{93.74} & \underline{0.47} \\
    toothbrush
      & 40.81 & 0.19 & 87.25 & \textbf{0.42}
      & \textbf{9.15} & 0.14 & \underline{10.64} & \underline{0.20}
      & 12.35 & 0.19 \\
    transistor
      & 116.43 & \textbf{0.32} & 140.03 & \underline{0.31}
      & \underline{38.03} & 0.29 & 67.90 & \textbf{0.32}
      & \textbf{29.16} & \textbf{0.32} \\
    wood
      & 84.71 & 0.35 & 71.51 & \textbf{0.38}
      & \textbf{23.98} & 0.35 & \underline{36.71} & 0.36
      & 41.70 & \underline{0.37} \\
    zipper
      & 164.87 & \underline{0.24} & 245.00 & \textbf{0.40}
      & \underline{57.73} & 0.21 & 62.34 & 0.23
      & \textbf{53.94} & 0.23 \\
    \midrule
    Avg.
      & \tableoneavgvalue{101.91} & \tableoneavgvalue{\underline{0.30}}
      & \tableoneavgvalue{114.73} & \tableoneavgvalue{\textbf{0.39}}
      & \tableoneavgvalue{\underline{32.38}} & \tableoneavgvalue{0.28}
      & \tableoneavgvalue{47.63} & \tableoneavgvalue{0.29}
      & \tableoneavgvalue{\textbf{31.09}} & \tableoneavgvalue{\underline{0.30}} \\
    \bottomrule
  \end{tabular*}

  \vspace{4pt}
  \par
  \centering
  \captionof{table}{Anomaly classification accuracy (\%) on MVTec-AD.}
  \label{tab:classification}
  \fontsize{7.3pt}{7.8pt}\selectfont
  \setlength{\tabcolsep}{1.7pt}
  \renewcommand{\arraystretch}{1.04}
  \begin{tabular*}{\columnwidth}{@{\extracolsep{\fill}}>{\raggedright\arraybackslash}p{0.17\columnwidth}|*{4}{>{\centering\arraybackslash}p{0.145\columnwidth}|}>{\centering\arraybackslash}p{0.145\columnwidth}@{}}
    \toprule
    Category & AnoDiff & DualAno & TF-IDG & O2MAG & Ours \\
    \midrule
    bottle     & 83.72 & 62.79 & \textbf{90.70} & \underline{88.37} & 86.05 \\
    cable      & \underline{65.63} & \textbf{73.44} & 57.81 & \textbf{73.44} & \underline{65.63} \\
    capsule    & 41.33 & \underline{56.00} & \underline{56.00} & 50.67 & \textbf{61.33} \\
    carpet     & 43.55 & 38.71 & \underline{80.65} & \underline{80.65} & \textbf{82.26} \\
    grid       & 35.00 & 50.00 & 55.00 & \textbf{62.50} & \underline{60.00} \\
    hazelnut   & 79.17 & 62.50 & \underline{91.67} & 89.58 & \textbf{95.83} \\
    leather    & 61.90 & 76.19 & \textbf{82.54} & \underline{80.95} & \textbf{82.54} \\
    metal\_nut & 73.44 & \textbf{95.31} & 67.19 & \underline{84.38} & 73.44 \\
    pill       & 23.96 & 32.29 & 37.50 & \textbf{48.96} & \underline{41.67} \\
    screw      & 34.57 & \underline{35.80} & 33.33 & \textbf{37.04} & \underline{35.80} \\
    tile       & \underline{89.47} & 87.72 & 82.46 & \textbf{100} & \textbf{100} \\
    transistor & 64.29 & \textbf{92.86} & 67.86 & \underline{82.14} & 71.43 \\
    wood       & \underline{73.81} & 66.67 & \textbf{78.57} & \textbf{78.57} & \underline{73.81} \\
    zipper     & 58.54 & 32.93 & 48.78 & \textbf{71.95} & \underline{60.98} \\
    \midrule
    Avg. & 59.17 & 61.66 & 66.43 & \textbf{73.51} & \underline{70.77} \\
    \bottomrule
  \end{tabular*}
\end{minipage}\hfill
\begin{minipage}[t]{0.48\textwidth}
  \vspace{0pt}
  \centering
  \includegraphics[width=\linewidth]{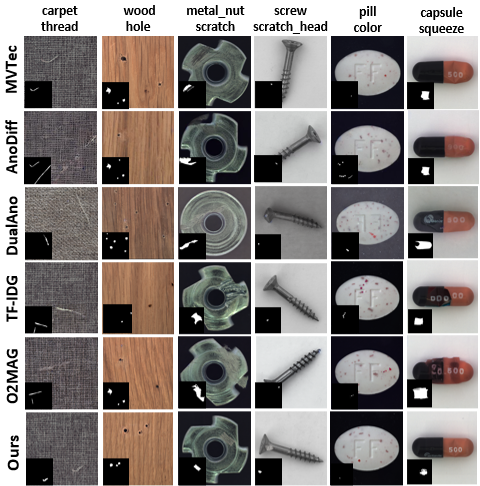}
  \vspace{-6pt}
  \refstepcounter{figure}
  \caption*{\textbf{Figure~\thefigure.} Qualitative comparison on MVTec-AD. Lower-left insets show the anomaly masks.}
  \label{fig:qualitative-comparison}
  \label{fig:qualitative}

  \vspace{-2pt}
  \setcounter{table}{3}
  \captionof{table}{Zero-shot comparison on Hazelnut.}
  \label{tab:zero-shot-comparison}
  \fontsize{6.8pt}{7.6pt}\selectfont
  \setlength{\tabcolsep}{0.6pt}
  \renewcommand{\arraystretch}{1.05}
  \begin{tabular*}{\linewidth}{@{\extracolsep{\fill}}lcccccc@{}}
    \toprule
    Method & KID$\downarrow$ & IC-L$\uparrow$ & AP-I & AUC-P & AP-P & F1-P \\
    \midrule
    AnoStyler & 96.48 & 0.07 & 94.63 & 97.17 & 66.02 & 61.94 \\
    \textbf{REDiff-AD} & \textbf{38.54} & \textbf{0.32} & \textbf{95.54} & \textbf{97.31} & \textbf{66.13} & \textbf{70.72} \\
    \bottomrule
  \end{tabular*}
  \setcounter{table}{2}
\end{minipage}
\par\vspace{-3pt}
\end{table*}%
}

\newcommand{\placeunettable}{%
\begin{table*}[!t]
  \centering
  \caption{Trained U-Net anomaly detection and localization on MVTec-AD. Higher is better.}
  \label{tab:unet-downstream}
  \label{tab:detection}
  \setlength{\tabcolsep}{1.15pt}
  \renewcommand{\arraystretch}{1.01}
  \resizebox{\textwidth}{!}{%
  \begin{tabular}{l|cccc|cccc|cccc|cccc|cccc}
    \toprule
    \multirow{2}{*}{Category}
      & \multicolumn{4}{c|}{AnomalyDiffusion}
      & \multicolumn{4}{c|}{DualAnoDiff}
      & \multicolumn{4}{c|}{TF-IDG}
      & \multicolumn{4}{c|}{O2MAG}
      & \multicolumn{4}{c}{Ours} \\
    \cline{2-21}
      & AP-I & AUC-P & AP-P & F1-P
      & AP-I & AUC-P & AP-P & F1-P
      & AP-I & AUC-P & AP-P & F1-P
      & AP-I & AUC-P & AP-P & F1-P
      & AP-I & AUC-P & AP-P & F1-P \\
    \midrule
    bottle
      & 99.5 & \textbf{99.7} & \textbf{94.9} & \textbf{87.1}
      & 99.2 & 98.9 & 90.3 & 83.3
      & \textbf{100} & \underline{99.4} & 91.6 & 84.2
      & \underline{99.8} & 99.3 & \underline{92.8} & \underline{85.0}
      & 99.6 & 98.8 & 89.7 & 83.7 \\
    cable
      & \underline{99.2} & \underline{98.4} & \underline{83.2} & \underline{75.5}
      & 89.6 & 94.1 & 63.2 & 60.2
      & 99.1 & 92.9 & 67.6 & 63.5
      & 99.0 & \textbf{98.6} & \textbf{85.5} & \textbf{77.9}
      & \textbf{99.3} & 95.1 & 73.9 & 70.1 \\
    capsule
      & \underline{96.5} & 93.9 & \underline{45.3} & 47.4
      & 93.5 & 95.0 & 38.2 & 47.9
      & 92.4 & \underline{95.6} & 40.2 & \textbf{51.0}
      & 96.2 & 95.3 & 44.7 & \underline{49.4}
      & \textbf{98.6} & \textbf{96.2} & \textbf{46.8} & 49.0 \\
    carpet
      & 98.2 & \textbf{99.2} & \textbf{83.9} & \underline{75.3}
      & 97.3 & 98.8 & 77.2 & 70.5
      & \underline{98.8} & \underline{99.1} & 83.2 & 75.1
      & 98.3 & 98.7 & \underline{83.7} & \textbf{75.9}
      & \textbf{99.1} & 98.6 & 80.8 & 73.4 \\
    grid
      & 98.2 & 92.9 & 29.8 & 36.3
      & 98.1 & 96.8 & \underline{55.9} & \underline{57.4}
      & 98.2 & 93.9 & 42.4 & 48.3
      & \textbf{100} & \underline{98.4} & 53.4 & 53.7
      & \underline{99.8} & \textbf{99.0} & \textbf{65.7} & \textbf{63.8} \\
    hazelnut
      & 99.1 & 99.3 & \underline{91.1} & \underline{85.1}
      & 98.9 & \textbf{99.6} & 90.8 & 83.8
      & \textbf{100} & 99.3 & 82.9 & 77.2
      & 97.4 & 99.3 & 82.5 & 76.3
      & \underline{99.7} & \underline{99.4} & \textbf{91.6} & \textbf{86.9} \\
    leather
      & \textbf{100} & \textbf{99.9} & 88.8 & 79.7
      & \textbf{100} & \textbf{99.9} & \textbf{90.8} & \textbf{82.8}
      & \textbf{100} & \underline{99.7} & 86.5 & 79.3
      & \textbf{100} & 99.3 & 81.8 & 76.1
      & \textbf{100} & \textbf{99.9} & \underline{90.4} & \underline{82.0} \\
    metal\_nut
      & \textbf{100} & \textbf{99.9} & \textbf{99.3} & \underline{95.8}
      & \textbf{100} & \textbf{99.9} & \textbf{99.3} & \textbf{95.9}
      & \textbf{100} & 99.6 & 98.5 & 94.7
      & \textbf{100} & \underline{99.8} & \underline{99.1} & 95.6
      & \underline{99.9} & \underline{99.8} & 98.9 & 95.0 \\
    pill
      & 99.3 & \underline{99.7} & \underline{95.5} & 88.3
      & 98.0 & 99.3 & 91.3 & 84.1
      & 98.4 & 99.6 & 93.8 & 86.9
      & \underline{99.4} & 99.6 & 95.2 & \underline{89.2}
      & \textbf{99.7} & \textbf{99.8} & \textbf{96.3} & \textbf{89.8} \\
    screw
      & 92.3 & 98.2 & \underline{50.6} & \underline{51.2}
      & 87.7 & 98.0 & 33.2 & 39.5
      & \underline{93.8} & 97.9 & 41.9 & 44.0
      & 92.3 & \textbf{99.0} & 49.3 & \textbf{52.3}
      & \textbf{96.8} & \underline{98.6} & \textbf{52.3} & \textbf{52.3} \\
    tile
      & \textbf{100} & \textbf{99.8} & \textbf{97.0} & \textbf{91.0}
      & 98.4 & 99.4 & 93.7 & 86.1
      & \underline{99.4} & \underline{99.6} & \underline{96.9} & \underline{90.6}
      & 98.9 & \underline{99.6} & 95.6 & 88.2
      & \textbf{100} & 99.4 & 95.7 & 89.2 \\
    toothbrush
      & \textbf{100} & \textbf{98.8} & \underline{75.5} & \underline{72.1}
      & \textbf{100} & \underline{98.7} & \textbf{80.3} & \textbf{74.3}
      & \textbf{100} & 98.3 & 57.6 & 58.2
      & \textbf{100} & 84.0 & 25.2 & 34.2
      & \textbf{100} & 93.7 & 43.5 & 49.7 \\
    transistor
      & \textbf{100} & \textbf{99.6} & \underline{92.9} & \underline{86.5}
      & 99.2 & 98.8 & 89.5 & 82.3
      & \underline{99.6} & 98.4 & 85.0 & 77.6
      & 97.3 & \underline{99.4} & \textbf{93.6} & \textbf{86.7}
      & \underline{99.6} & 95.6 & 78.4 & 72.4 \\
    wood
      & \textbf{99.8} & \underline{99.3} & \underline{85.1} & 76.6
      & 98.7 & \textbf{99.4} & \textbf{87.8} & \textbf{80.0}
      & 99.2 & 98.0 & 76.8 & 70.7
      & \textbf{99.8} & 99.2 & 83.1 & 75.1
      & \underline{99.6} & 99.0 & 83.9 & \underline{77.3} \\
    zipper
      & \textbf{100} & \textbf{99.6} & \textbf{88.3} & \textbf{80.4}
      & \textbf{100} & \underline{99.2} & 81.0 & 73.8
      & \textbf{100} & 98.4 & \underline{86.5} & \underline{80.2}
      & \textbf{100} & \underline{99.2} & 84.0 & 76.2
      & \textbf{100} & 97.6 & 79.4 & 76.2 \\
    \midrule
    Avg.
      & \underline{98.8} & \textbf{98.5} & \textbf{80.1} & \textbf{75.2}
      & 97.2 & \underline{98.4} & 77.5 & 73.5
      & 98.6 & 98.0 & 75.4 & 72.1
      & 98.6 & 97.9 & 76.6 & 72.8
      & \textbf{99.4} & 98.0 & \underline{77.8} & \underline{74.1} \\
    \bottomrule
  \end{tabular}%
  }
\end{table*}%

\setcounter{table}{4}
\setcounter{topnumber}{1}
\begin{figure}[t]
  \centering
  \includegraphics[width=\linewidth]{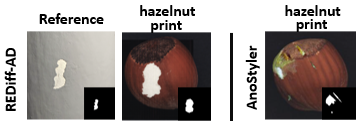}
  \vspace{-8pt}
  \refstepcounter{figure}
  \caption*{\textbf{Figure~\thefigure.} Zero-shot comparison on Hazelnut print. REDiff-AD uses an ordinary visual reference; AnoStyler is prompt-driven.}
  \label{fig:zero-shot-qualitative}

  \vspace{2pt}
  \captionof{table}{Efficiency comparison of training-free anomaly generation methods on an NVIDIA RTX~5090.}
  \label{tab:efficiency}
  \fontsize{6.8pt}{7.2pt}\selectfont
  \setlength{\tabcolsep}{1.2pt}
  \renewcommand{\arraystretch}{0.94}
  \begin{tabular*}{\linewidth}{@{\extracolsep{\fill}}lccc@{}}
    \toprule
    Method & Time (s / image) & Max VRAM & KID$\downarrow$ \\
    \midrule
    TF-IDG~\cite{xu2025tfidg} & 22~s & 22~GB & \underline{8.62} \\
    O2MAG~\cite{rao2026o2mag} & 20~s & \underline{16~GB} & 9.74 \\
    \midrule
    REDiff-AD (20 steps) & \textbf{10~s} & \textbf{9~GB} & 9.26 \\
    REDiff-AD (30 steps) & \underline{16~s} & \textbf{9~GB} & 8.87 \\
    REDiff-AD (40 steps) & 22~s & \textbf{9~GB} & \textbf{8.53} \\
    \bottomrule
  \end{tabular*}
  \vspace{-2pt}
\end{figure}%

\begin{table}[t]
  \centering
  \caption{Mask-generation ablation on MVTec-AD.}
  \label{tab:mask-ablation}
  \fontsize{7.2pt}{7.7pt}\selectfont
  \setlength{\tabcolsep}{2.2pt}
  \renewcommand{\arraystretch}{0.96}
  \begin{tabular*}{\linewidth}{@{\extracolsep{\fill}}l!{\hspace{18pt}}|cccc@{}}
    \toprule
    \tablefiveheader{Metric} & \tablefiveheader{ROI}
      & \tablefiveheader{\makecell{No block\\selection}}
      & \tablefiveheader{\makecell{Selected\\blocks}}
      & \tablefiveheader{\makecell{Refined\\mask}} \\
    \midrule
    AUC-P & 95.43 & 96.48 & \underline{97.15} & \textbf{98.03} \\
    AP-P  & 69.98 & 70.16 & \underline{72.82} & \textbf{77.82} \\
    F1-P  & 67.73 & 67.85 & \underline{70.24} & \textbf{74.05} \\
    \bottomrule
  \end{tabular*}

  \vspace{5pt}
  \caption{Component ablation on MVTec-AD, $\checkmark$ denotes enabled.}
  \label{tab:component-ablation}
  \fontsize{7pt}{7.4pt}\selectfont
  \setlength{\tabcolsep}{1.0pt}
  \renewcommand{\arraystretch}{0.92}
  \begin{tabular*}{\linewidth}{@{\extracolsep{\fill}}ccc!{\hspace{12pt}}|cccc@{}}
    \toprule
    \multicolumn{3}{c!{\hspace{12pt}}|}{Components} & \multicolumn{4}{c}{Metrics} \\
    In-context & AAM & DOK
      & KID$\downarrow$ & IC-L$\uparrow$ & Acc. & AP-I \\
    \midrule
    $\checkmark$ &              &              & 35.62 & \underline{0.28} & 61.07 & 97.5 \\
    $\checkmark$ & $\checkmark$ &              & \textbf{28.92} & 0.27 & \underline{67.91} & \underline{98.6} \\
    $\checkmark$ &              & $\checkmark$ & 32.04 & \textbf{0.30} & 63.31 & 98.1 \\
    $\checkmark$ & $\checkmark$ & $\checkmark$
      & \underline{31.09} & \textbf{0.30} & \textbf{70.77} & \textbf{99.4} \\
    \bottomrule
  \end{tabular*}
\end{table}%
}

%% file: sections/conclusion.tex
\section{Conclusion}
\label{sec:conclusion}
We proposed REDiff-AD, a simple and efficient framework for industrial anomaly generation without additional model training or manually specified target masks. By introducing in-context defect transfer into a modern Flow Matching MMDiT, REDiff-AD brings the latest generative architecture to industrial anomaly synthesis, enabling realistic reference-guided generation with efficient sampling. Internal attention is further reused to produce aligned pixel-level pseudo-masks and provide adaptive feedback for robust defect generation, while DOK addresses the limitations of in-context generation to improve anomaly diversity. Experiments show that REDiff-AD achieves superior performance in the generation quality and downstream classification, detection, and localization tasks, while requiring lower computational and data-preparation costs.

%% file: sections/acknowledgments.tex
\section*{Acknowledgments}

This work was supported in part by the National Science and Technology Council (NSTC), Taiwan. We sincerely thank our advisor, Prof.~Jian-Jiun Ding, for the valuable guidance and support throughout this research. We also thank the members of the DISP Lab at National Taiwan University for their helpful discussions, assistance, and encouragement.
\par